\documentclass[journal]{IEEEtran}
\IEEEoverridecommandlockouts
\usepackage{cite}
\usepackage{amsmath,amssymb}
\usepackage{graphicx}
\usepackage[dvipsnames]{xcolor}
\usepackage{url}
\usepackage{booktabs}
\definecolor{cvprblue}{rgb}{0.21,0.49,0.74}
\usepackage[breaklinks,colorlinks,allcolors=cvprblue]{hyperref}
\makeatletter
\let\mir@maketitle\@maketitle
\renewcommand{\@maketitle}{\mir@maketitle
\vspace{10pt}
\begin{center}\includegraphics[width=\textwidth]{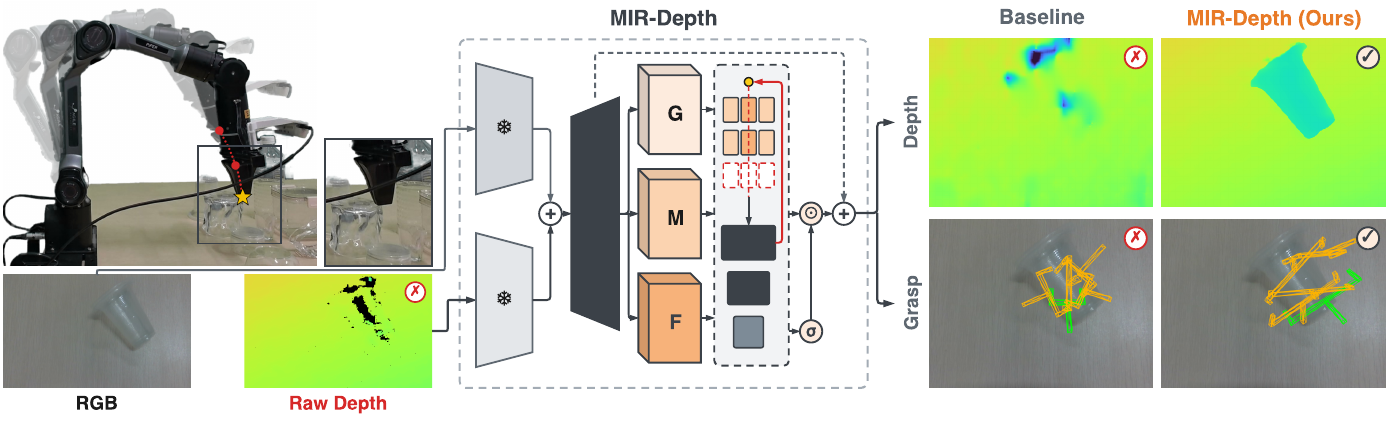}\end{center}
\vspace{-6pt}{\footnotesize\setcounter{figure}{0}\refstepcounter{figure}\label{fig:teaser}Fig.~\thefigure.\enskip \textbf{Single-view depth completion of transparent objects for robotic grasping.} Given a transparent cup captured by a wrist-mounted RGB-D camera (\textit{left}), raw sensor depth suffers from specular holes and refracted background readings. Benefiting from a prior-guided initial anchor, \textbf{MIR-Depth} (\textit{middle}) pairs one shared global and two pixel-wise local inverse-depth volumes with validity-masked 1D lookup and spatial gating, recovering accurate visible surfaces over single-pass baselines for 6-DoF grasp pose prediction (\textit{right}).\par}
\vspace{8pt}}
\makeatother

\begin{document}
\title{MIR-Depth: Multi-Range Refinement for Single-View RGB-D Depth Completion of Transparent Objects}
\author{Xinzhe Fu$^{*}$, Qi Shao$^{*\dagger}$, Yuanfei Xie$^{*}$, Taoyu Wu, Zhihao Cao, Long Hu, Hong Liang, Xianzhi Li, Anh Nguyen, Baoru Huang, and Rongsheng Li%
\thanks{$^{*}$X. Fu, Q. Shao, and Y. Xie contributed equally to this work. $^{\dagger}$Q. Shao is the project leader. X. Fu developed the algorithm, and Y. Xie deployed the system on the real robot.}%
\thanks{X. Fu, Q. Shao, Y. Xie, H. Liang, and R. Li are with Harbin Engineering University, Harbin, China. Q. Shao is also with the University of Liverpool, Liverpool, UK (e-mail: shaoqi@liverpool.ac.uk).}%
\thanks{T. Wu is with the University of Liverpool, Liverpool, UK, and also with Xi'an Jiaotong-Liverpool University, Suzhou, China.}%
\thanks{Z. Cao is with ETH Zurich, Zurich, Switzerland.}%
\thanks{L. Hu and X. Li are with Huazhong University of Science and Technology, Wuhan, China.}%
\thanks{A. Nguyen and B. Huang are with the University of Liverpool, Liverpool, UK, and also with Imperial College London, London, UK.}%
}
\maketitle

\begin{abstract}
Recovering accurate visible-surface depth from a single RGB-D image and an object mask is essential for robotic grasping and manipulation of transparent objects.
Specular reflections cause missing depth, while refraction corrupts nonzero returns with background depth, making raw measurements unreliable as surface anchors.
Under a fixed sampling budget, refining an initial single-view prediction within a single search interval either misses large refraction offsets or sacrifices fine local resolution.
To resolve this single-view trade-off, MIR-Depth constructs Coordinate-Aware Multi-Range Hypotheses: a 3D-aggregated global inverse-depth volume and depth-axis-filtered pixel-wise local volumes around a prior-guided initial depth.
Selective Recurrent Refinement then introduces out-of-bounds lookup masking and validity-masked softmax fusion to iteratively update inverse depth from these volumes using a multi-scale ConvGRU.
Finally, Spatially Gated Correction applies a learned per-pixel spatial gate to the full-resolution residual, preserving accurate initial geometry without imposing raw sensor readings as hard constraints.
Experiments on TransCG, TODD, TRansPose, ClearPose, and HouseCat6D show that MIR-Depth achieves lower error than ReMake, DFNet, BP-Net, and PSD. Real-robot evaluations confirm accurate surface geometry for 6-DoF grasp pose prediction.
Code and pretrained models will be publicly available.

\end{abstract}

\section{Introduction}
\label{sec:intro}

Transparent objects are common in household and laboratory settings, yet recovering their surface geometry remains challenging for robotic manipulation (Fig.~\ref{fig:teaser}).
Accurate visible-surface depth supports grasp planning~\cite{fang2020graspnet}, but reflection and refraction cause missing or erroneous nonzero RGB-D measurements that may correspond to background surfaces~\cite{sajjan2020clear,liu2020keypose,zhu2021rgb}.
Opaque-scene depth completion often uses observed depths as geometric anchors~\cite{cheng2018depth,park2020non,zhang2023completionformer}, whereas transparent-region readings may lie behind the surface a gripper must contact~\cite{cheng2026rethinking}.
An object mask localizes the target but cannot establish the reliability of nonzero readings, so single-view completion must fill missing depth and correct corrupted measurements without treating them as reliable surface constraints.


Single-view networks predict dense depth from RGB-D inputs in one forward pass~\cite{fang2022transcg,chen2022tode,li2023fdct}, while prior-guided methods incorporate monocular depth estimates~\cite{ranftl2020towards,yang2024depth,yang2024depthv2}, as in ReMake, which combines a relative-depth prior with an object mask~\cite{cheng2026rethinking}.
Despite providing a useful metric reference, prior-guided predictions can have residual errors ranging from a few millimeters along thin rims to tens of centimeters in strongly refracted interiors.
With a fixed per-pixel sampling budget, narrow search intervals miss large refraction offsets, whereas workspace-wide intervals reduce local sampling resolution.
Multi-range disparity volumes address a similar trade-off in stereo matching~\cite{gu2020cascade,xu2025igevpp}, but single-view RGB-D completion lacks epipolar correlation and must score candidates using monocular context and unreliable raw depth.

MIR-Depth (Fig.~\ref{fig:overview}) iteratively refines single-view depth across multiple inverse-depth volumes using metric initialization and multi-scale context from RGB, raw depth, the object mask, and a frozen Depth Anything V2 prior~\cite{yang2024depthv2}.
Coordinate-Aware Multi-Range Hypotheses combines one shared global volume with two pixel-wise local volumes around the initial reference inverse depth, using validity-masked raw-depth residuals as soft evidence.
Since global bins share scene-wide planes while local bins shift per pixel, a 3D hourglass aggregates the global volume spatially, whereas depth-axis convolutions filter local volumes without cross-pixel depth mixing.
Selective Recurrent Refinement uses out-of-bounds lookup masking and validity-masked softmax fusion to combine fine local detail and wide global corrections through a multi-scale ConvGRU.
Spatially Gated Correction applies a learned per-pixel gate to the full-resolution residual, preserving accurate initial regions without clamping to raw depth.
Together, these designs reduce depth error on TransCG and four unseen zero-shot benchmarks while supporting reliable 6-DoF grasping on a physical robot.

Our main contributions are:
\begin{itemize}
    \item We propose Coordinate-Aware Multi-Range Hypotheses, pairing a 3D-aggregated global inverse-depth volume with depth-axis-filtered pixel-wise local volumes to cover large refraction offsets and fine surface detail.
    \item We design Selective Recurrent Refinement, introducing out-of-bounds lookup masking and validity-masked softmax fusion to iteratively update single-view inverse depth from monocular context and soft raw-depth residuals.
    \item We introduce Spatially Gated Correction, learning a full-resolution per-pixel spatial gate to blend recurrent depth residuals into the initial estimate without treating nonzero raw depth as a hard constraint.
\end{itemize}


\section{Related Work}
\label{sec:related_work}

\subsection{Transparent Object Depth Completion}

Conventional RGB-D and LiDAR completion methods recover dense depth through spatial propagation~\cite{cheng2018depth,park2020non} or learned feature fusion~\cite{zhang2023completionformer,tang2024bilateral,chen2025propagating}.
For transparent objects, reflection and refraction cause missing depth and erroneous nonzero measurements that may correspond to background surfaces, requiring both hole filling and correction of unreliable observations~\cite{sajjan2020clear,liu2020keypose}.

ClearGrasp~\cite{sajjan2020clear} uses predicted surface normals and occlusion boundaries to guide depth optimization.
Other methods learn local implicit representations over ray--voxel pairs~\cite{zhu2021rgb} or jointly complete point clouds and depth maps~\cite{xu2022seeingglass}.
RGB-D regression networks~\cite{dai2022domain,fang2022transcg,hong2022cluedepth,chen2022tode,li2023fdct} predict dense depth from image and sensor features.
This line of work uses synthetic data~\cite{dai2022domain} and real-world datasets~\cite{fang2022transcg,chen2022clearpose,kim2024transpose,jung2024housecat6d} for training and evaluation, with completed geometry supporting 6-DoF grasp detection~\cite{fang2020graspnet}.

Stereo-based methods~\cite{shi2024asgrasp,min2026depthfocus} use cross-view observations, while monocular multi-layer estimation~\cite{wen2026seegroup} predicts multiple surfaces along each viewing ray.
For single-view completion, pretrained monocular depth estimators~\cite{ranftl2020towards,yang2024depth,yang2024depthv2} provide scene priors that complement raw depth.
ReMake~\cite{cheng2026rethinking} combines a relative-depth prior and an object mask for RGB-D depth prediction, while MIR-Depth refines a prior-guided initial estimate to recover visible-surface depth from a single RGB-D image and object mask.

\subsection{Iterative Depth Refinement}

PSMNet~\cite{chang2018pyramid} aggregates disparity cost volumes with stacked 3D hourglass networks, while CasMVSNet~\cite{gu2020cascade} progressively narrows depth intervals in a coarse-to-fine cascade.
Building on RAFT's recurrent lookup framework~\cite{teed2020raft}, RAFT-Stereo~\cite{lipson2021raft} and IGEV-Stereo~\cite{xu2023iterative} update disparity through local volume queries and multi-level ConvGRUs.
IGEV++~\cite{xu2025igevpp} extends this approach with multiple disparity ranges to retain fine detail while covering large disparities.
These methods derive candidate evidence from cross-view feature matching, unavailable in single-view RGB-D completion.

In monocular depth estimation, IEBins~\cite{shao2023iebins} progressively adjusts depth bins based on estimated depth and uncertainty.
For transparent-object completion, LIDF~\cite{zhu2021rgb} refines its initial prediction through self-correcting updates, DCIRNet~\cite{xie2026dcirnet} combines refinement with multi-stage supervision, and TransDiff~\cite{wang2025transdiff} uses iterative diffusion denoising conditioned on RGB-D information.

With a fixed sampling budget, candidate-based refinement trades off large-error coverage and fine local resolution.
MIR-Depth combines a shared global inverse-depth range with two pixel-wise local ranges around a prior-guided reference for single-view transparent-object completion.
It scores candidates using image features and raw-depth residuals, adapting volume aggregation and recurrent lookup to shared global and pixel-dependent local coordinates.

\begin{figure*}[t]
    \centering
    \includegraphics[width=0.98\textwidth]{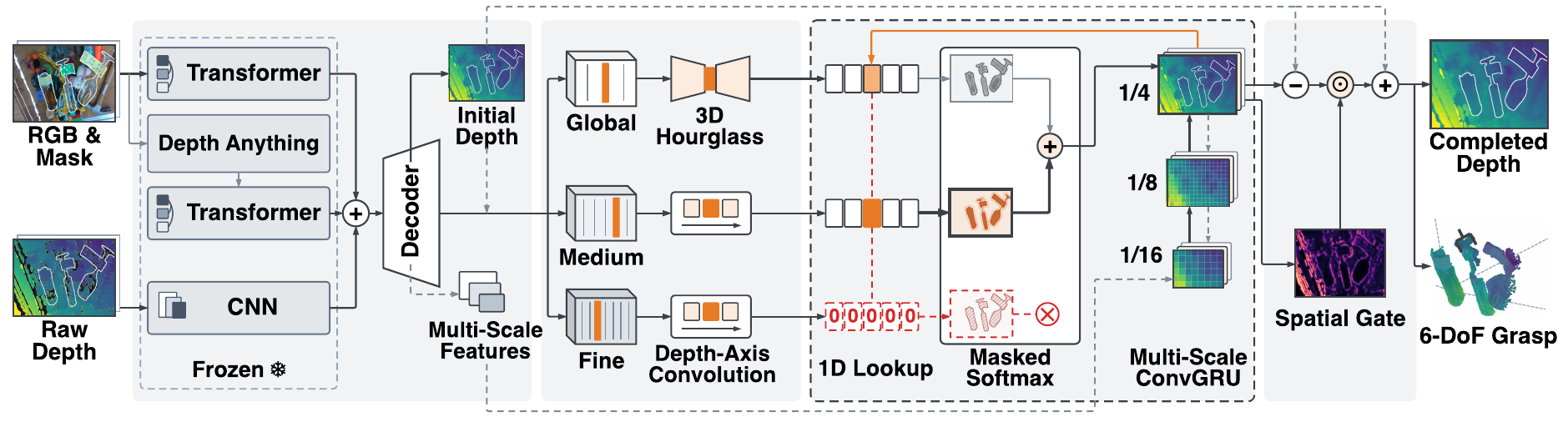}
    \vspace{-0.45em}
    \caption{\textbf{Overview of MIR-Depth.} A frozen Depth Anything V2 model provides the relative-depth prior. Transformer and CNN encoders extract RGB-mask, relative-depth, and raw-depth features, which are summed at each scale. The summed features provide multi-scale context, while a decoder and depth head predict initial depth $D_0$. Coordinate-Aware Multi-Range Hypotheses construct one shared global inverse-depth volume aggregated by a 3D hourglass alongside medium and fine pixel-wise local volumes filtered by depth-axis convolutions. Selective Recurrent Refinement queries these volumes, encodes lookup features together with per-offset validity indicators, and fuses active ranges to update inverse depth through a multi-scale ConvGRU. Spatially Gated Correction weights the full-resolution residual relative to $D_0$ to produce completed depth for 6-DoF grasp pose prediction.}
    \label{fig:overview}
    \vspace{-0.35em}
\end{figure*}

\section{Method}
\label{sec:method}

Given an RGB image $I\in\mathbb{R}^{3\times H\times W}$, aligned raw depth $D_{\mathrm{raw}}\in\mathbb{R}^{1\times H\times W}$ in meters, and an object mask $M\in\{0,1\}^{1\times H\times W}$, MIR-Depth predicts visible-surface depth $\widehat D\in\mathbb{R}^{1\times H\times W}$, omitting batch indices throughout.
As shown in Fig.~\ref{fig:overview}, MIR-Depth predicts initial depth and multi-scale features, builds global and pixel-wise local candidate volumes with raw depth as soft evidence, refines inverse depth through valid-range lookup fusion, and gates the final residual.

\noindent\textbf{Prior-guided initialization.}
Let $D_{\mathrm{raw}}=0$ indicate missing depth and $V_{\mathrm{raw}}=\mathbf{1}[D_{\mathrm{raw}}>0]$ mark nonzero returns.
A frozen Depth Anything V2 model~\cite{yang2024depthv2} produces a relative-depth prior $R=f_{\mathrm{DAv2-S}}(I)\in\mathbb{R}^{1\times H\times W}$. Following ReMake~\cite{cheng2026rethinking}, the initializer combines this prior with RGB, raw depth, and the object mask:
\begin{equation}
    (D_0,F_4,F_8,F_{16})=f_{\mathrm{init}}(I,D_{\mathrm{raw}},R,M),
    \label{eq:base_prediction}
\end{equation}
where $D_0\in\mathbb{R}_{\ge 0}^{1\times H\times W}$ is the initial metric depth for candidate construction and residual correction. Context features $F_s\in\mathbb{R}^{C_s\times H/s\times W/s}$, $s\in\{4,8,16\}$, sum corresponding encoder features; the decoder and depth head produce $D_0$.

\subsection{Coordinate-Aware Multi-Range Hypotheses}
\label{sec:method_hypotheses}

Coordinate-Aware Multi-Range Hypotheses combine a shared global range $g$ with medium and fine pixel-wise local ranges $m$ and $f$ constructed from $D_0$.
We resize $D_0$ to obtain $B$ and downsample $D_{\mathrm{raw}}$ and $V_{\mathrm{raw}}$ to obtain $d_{\mathrm{raw}}$ and $v_{\mathrm{raw}}$ on the same working grid; nearest-neighbor sampling of the raw inputs avoids averaging missing and nonzero measurements.
At pixel $p$, the reference inverse depth is $\bar q(p)=\operatorname{sg}[\mathcal Q(B(p))]$, where $\mathcal Q(B)=\operatorname{clip}(1/\max(B,\epsilon),q_-,q_+)$, $\epsilon>0$ is a depth floor, and $0<q_-<q_+$ bound the allowed inverse depth. The operator $\operatorname{sg}[\cdot]$ stops gradients through the reference; $p$ is omitted below when clear.


\noindent\textbf{Candidate ranges.}
For each range $r\in\{g,m,f\}$, we uniformly sample $K\geq2$ inverse-depth candidates on $[a_r(p),b_r(p)]$ with $a_r(p)<b_r(p)$:
\begin{equation}
    q_{r,k}(p)=a_r(p)+\frac{k}{K-1}\bigl(b_r(p)-a_r(p)\bigr),
    \label{eq:candidate_grid}
\end{equation}
for $k\in\{0,\ldots,K-1\}$.
The global bounds $(a_g,b_g)$ are shared across pixels and lie within the allowed inverse-depth range: $q_-<a_g<b_g<q_+$. For $r\in\{m,f\}$, local bounds are $a_r(p)=\operatorname{clip}((1-\rho_r)\bar q(p),q_-,q_+)$ and $b_r(p)=\operatorname{clip}((1+\rho_r)\bar q(p),q_-,q_+)$, where $0<\rho_f<\rho_m<1$ controls their relative widths. All candidate grids remain fixed during recurrent refinement.
We set $m_g(p)=1$ and $m_m(p)=m_f(p)=\mathbf{1}[B(p)>0]$, invalidating local ranges at zero initial depth while retaining global candidates.

\noindent\textbf{Candidate evidence.}
Without stereo feature correlation, we combine pointwise context features $\phi$ projected from $F_4$ and resized $I,R$ with geometric compatibility cues.
For candidate inverse depth $q$ with $d=1/q$ and validity-masked raw residual $\delta_{\mathrm{obs}}=v_{\mathrm{raw}}(d-d_{\mathrm{raw}})$, we encode absolute depth, deviation from the initial estimate, and raw-depth agreement as

\begin{equation}
\begin{split}
    c(q,p)=\mathcal N_c\!\Big([&q,\ d,\ q-\bar q,\ d-B,\\
               &\delta_{\mathrm{obs}},\ |\delta_{\mathrm{obs}}|,\ v_{\mathrm{raw}},\ \mathbf{1}[B>0]]\Big),
\end{split}
\label{eq:candidate_conditions}
\end{equation}
where $\mathcal N_c$ divides inverse-depth components $q,q-\bar q$ by $s_q>0$ and depth components $d,d-B,\delta_{\mathrm{obs}},|\delta_{\mathrm{obs}}|$ by $s_d>0$, preserves validity indicators, and clips each component to $[-c_{\max},c_{\max}]$, with $c_{\max}\geq1$.
Range-specific pointwise projection encodes $[\phi(p),c(q_{r,k}(p),p)]$ into candidate volume $C_r$, using raw-depth agreement as soft evidence without constraining the predicted surface.


\noindent\textbf{Coordinate-aware aggregation.}
In the global volume $C_g$, bins share inverse depths across pixels, enabling joint spatial and inverse-depth aggregation into $A_g$~\cite{chang2018pyramid}. Local bins depend on $\bar q(p)$ and can represent different depths at neighboring pixels, so $C_m$ and $C_f$ are filtered only along the inverse-depth axis to obtain $A_m$ and $A_f$.
A pointwise scoring head $\psi_r$ then produces
\begin{equation}
    P_r(p,k)=m_r(p)\,\operatorname{softmax}_k\bigl(\psi_r(A_r(p,k))\bigr),
    \label{eq:candidate_probability}
\end{equation}
summing to one when $m_r(p)=1$ and zero otherwise.
For downstream range selection, we compute normalized entropy $H_r(p)=-\sum_k P_r(p,k)\log P_r(p,k)/\log K$ using $0\log 0=0$ and peak probability $\pi_r(p)=\max_k P_r(p,k)$, setting both to zero when $m_r(p)=0$.

\subsection{Selective Recurrent Refinement}
\label{sec:method_updates}

Using $(A_r,P_r)$, we refine inverse-depth query $q^t(p)$ over $t=1,\ldots,T$ steps, initializing $q^0(p)=\bar q(p)$ when $B(p)>0$ and using global expectation $q^0(p)=\sum_k P_g(p,k)q_{g,k}$ otherwise.

\noindent\textbf{Coordinate-consistent lookup.}
At step $t$, query $q^{t-1}(p)$ maps to $x_r^t(p)=(K-1)(q^{t-1}(p)-a_r(p))/(b_r(p)-a_r(p))$ in each range. Following recurrent lookup~\cite{teed2020raft,lipson2021raft}, we linearly interpolate $[A_r,P_r]$ at $x_r^t(p)+\Delta$ for offsets $\Delta\in\mathcal D=\{-r_l,\ldots,r_l\}$, where $r_l$ is a nonnegative integer lookup radius.
Clamping a query outside a local interval would return edge-bin features for a depth beyond that interval.
An offset has validity indicator $v_{r,\Delta}^t(p)=1$ only when $m_r(p)=1$, $q^{t-1}(p)\in[a_r(p),b_r(p)]$, and $x_r^t(p)+\Delta\in[0,K-1]$; otherwise $v_{r,\Delta}^t(p)=0$. Invalid samples are zeroed, and sampled features are concatenated over $\mathcal D$ into $L_r^t(p)$.

\noindent\textbf{Fusion over valid ranges.}
Each range-specific encoder maps $L_r^t$ and per-offset indicators $\{v_{r,\Delta}^t\}_{\Delta\in\mathcal D}$ to $e_r^t$, with $e_r^t=0$ where no offset is valid. A learned pointwise head uses $[e_r^t,H_r,\pi_r,v_{\mathrm{raw}}]$ to predict selection logits $\ell_r^t(p)$.
Let $\mathcal S_t(p)$ contain ranges with at least one valid lookup offset. For $\mathcal S_t(p)\neq\emptyset$ and $r\in\mathcal S_t(p)$, validity-masked softmax fusion gives
\begin{equation}
    \alpha_r^t=\frac{\exp(\ell_r^t)}{\sum_{j\in\mathcal S_t}\exp(\ell_j^t)},
    \qquad e^t=\sum_{r\in\mathcal S_t}\alpha_r^t e_r^t,
    \label{eq:selective_fusion}
\end{equation}
with $\alpha_r^t(p)=0$ outside $\mathcal S_t(p)$ and $e^t(p)=0$ if $\mathcal S_t(p)=\emptyset$.

\noindent\textbf{Inverse-depth updates.}
A multi-scale ConvGRU, initialized and conditioned by $F_4,F_8,F_{16}$, combines spatial context by updating hidden states in the order $1/16$, $1/8$, and $1/4$ of input resolution.
At step $t$, the normalized query state is $s^t=\mathcal N_s([q^{t-1},\bar q,d_{\mathrm{raw}},v_{\mathrm{raw}}])$, where $\mathcal N_s$ divides the two inverse-depth components by $s_q$ and raw depth by $s_d$, retains the validity indicator, and clips all components to $[-c_{\max},c_{\max}]$. Using $s^t$, fused embedding $e^t$, and $\{\alpha_r^t,H_r,\pi_r\}_{r\in\{g,m,f\}}$, the updater predicts $u^t$ for the multiplicative inverse-depth update:
\begin{equation}
    q^t=\operatorname{clip}\!\left(
    q^{t-1}\exp\bigl(\eta\tanh(u^t)\bigr),q_-,q_+\right),
    \label{eq:inverse_update}
\end{equation}
where $\eta>0$ controls step magnitude while preserving positive inverse depth within $[q_-,q_+]$; upsampling $\mathcal U$ to input resolution yields $D^t=1/\mathcal U(q^t)$ at each step.

\subsection{Spatially Gated Correction}
\label{sec:method_gate}

To limit changes where $D_0$ is already accurate, Spatially Gated Correction predicts a per-pixel weight for the residual $D^T-D_0$.
The gate head uses the final recurrent hidden state, last lookup embedding $e^T$, summaries $\{H_r,\pi_r\}_{r\in\{g,m,f\}}$, and last pre-update query state $s^T$, which contains $q^{T-1}$. Upsampled logits pass through sigmoid $\sigma$ to form a per-pixel gate $G\in[0,1]^{1\times H\times W}$:
\begin{equation}
    \widehat D=D_0+G\odot(D^T-D_0),
    \label{eq:gated_prediction}
\end{equation}
with $\odot$ denoting element-wise multiplication.
The weights $\alpha_r^t$ combine valid-range features during lookup, while $G$ interpolates between $D_0$ and $D^T$ per pixel. The final correction applies across the image without masking by $M$ or clamping to $D_{\mathrm{raw}}$.

\subsection{Training Objective}
\label{sec:method_training}

The objective combines final and intermediate supervision with correction regularization relative to $D_0$.
For ground-truth depth $D^*$ and a binary target-validity mask $V^*$, define $\Omega=\{p:V^*(p)=1,\ D^*(p)>0\}$, $\Omega_M=\{p\in\Omega:M(p)=1\}$, and $\langle X\rangle_S=\sum_{p\in S}X(p)/\max(1,|S|)$.
The reconstruction loss combines depth errors on $\Omega$ and $\Omega_M$ with a gradient loss:
\begin{equation}
    \mathcal L_{\mathrm{rec}}=\langle|\widehat D-D^*|\rangle_\Omega
    +\langle|\widehat D-D^*|\rangle_{\Omega_M}+\lambda_{\mathrm{grad}}\mathcal L_{\mathrm{grad}},
    \label{eq:reconstruction_loss}
\end{equation}
Let $\mathcal E$ contain each horizontal or vertical adjacent pixel pair with both endpoints in $\Omega$, counted once. The gradient loss is
\begin{equation}
\mathcal L_{\mathrm{grad}}=
\frac{\sum_{(p,p')\in\mathcal E}
\left|\bigl(\widehat D(p')-\widehat D(p)\bigr)-\bigl(D^*(p')-D^*(p)\bigr)\right|}
{\max(1,|\mathcal E|)}.
\label{eq:gradient_loss}
\end{equation}
For $T\geq2$, the sequence loss supervises $D^1,\ldots,D^{T-1}$ with exponentially increasing weights controlled by $\gamma\in(0,1)$:
\begin{equation}
    \mathcal L_{\mathrm{seq}}=
    \frac{\sum_{t=1}^{T-1}\gamma^{T-1-t}\langle|D^t-D^*|\rangle_\Omega}
    {\sum_{t=1}^{T-1}\gamma^{T-1-t}}.
    \label{eq:sequence_loss}
\end{equation}
The correction regularizer $\mathcal L_{\mathrm{corr}}=\langle|\widehat D-D_0|\rangle_\Omega$ penalizes the magnitude of the final depth correction. The total loss is
\begin{equation}
    \mathcal L=\mathcal L_{\mathrm{rec}}
    +\lambda_{\mathrm{seq}}\mathcal L_{\mathrm{seq}}+\lambda_{\mathrm{corr}}\mathcal L_{\mathrm{corr}}.
    \label{eq:total_loss}
\end{equation}
Losses are computed per image and averaged across the batch. Valid ground-truth targets outside the candidate bounds remain included.


\section{Experiments}
\label{sec:experiments}

We evaluate MIR-Depth on TransCG~\cite{fang2022transcg}, zero-shot transfer to TODD~\cite{xu2022seeingglass}, TRansPose~\cite{kim2024transpose}, ClearPose~\cite{chen2022clearpose}, and HouseCat6D~\cite{jung2024housecat6d}, real-world 6-DoF grasp pose prediction, and component ablations. MAE and RMSE are reported in millimeters, $\mathrm{REL}$ is dimensionless, and $\delta_\tau$ is in percent.

\subsection{Implementation Details}
\label{sec:exp_setup}

We train MIR-Depth on the official TransCG split~\cite{fang2022transcg}, containing 34,191 single-view $640\times480$ RGB-D frames from 78 scenes, using RGB, raw depth, precomputed Depth Anything V2 Small priors, and dataset-provided transparent-object masks. The masks serve as input $M$ and define $\Omega_M$; processed training targets use $V^*(p)=\mathbf{1}[D^*(p)\geq0.01\,\mathrm{m}]$.
Following ReMake~\cite{cheng2026rethinking}, features from two Transformer encoders and one CNN encoder are summed at each scale and passed to a pyramid decoder and depth head. At quarter resolution, a three-level 3D hourglass aggregates the global volume, depth-axis convolutions filter local volumes, and a multi-scale ConvGRU updates inverse depth before spatial gating.
Raw depth and validity masks use nearest-neighbor downsampling, lookup uses linear interpolation, and $D_0$ resizing and inverse-depth and gate-logit upsampling use bilinear interpolation.

Feature widths are $(C_4,C_8,C_{16})=(48,96,192)$ for $(F_4,F_8,F_{16})$, $C_\phi=4$ for $\phi$, and $C_v=8$ for candidate volumes. We use $K=24$ candidates, $T=8$ iterations, and lookup offsets $\mathcal D=\{-r_l,\ldots,r_l\}$ with $r_l=2$.
We set $\epsilon=1/q_+=0.01\,\mathrm{m}$, $[q_-,q_+]=[0.05,100]\,\mathrm{m}^{-1}$, $(a_g,b_g)=(0.5,10)\,\mathrm{m}^{-1}$, and $(\rho_m,\rho_f)=(0.4,0.1)$.
Normalization scales are $(s_q,s_d)=(10\,\mathrm{m}^{-1},2\,\mathrm{m})$, with component-wise clipping at $c_{\max}=5$. We use $\eta=\log 2$, giving pre-clipping update factors in $[1/2,2]$, and $(\lambda_{\mathrm{grad}},\gamma,\lambda_{\mathrm{seq}},\lambda_{\mathrm{corr}})=(0.1,0.9,0.5,0.001)$.

The gate head is initialized with zero weights and bias $-4$, giving $G\approx0.018$ to limit initial changes to $D_0$.
We train refinement and gating for 5 epochs with $f_{\mathrm{init}}$ frozen, then jointly fine-tune them with the initializer decoder and depth head for 15 epochs, keeping the encoders, batch-normalization statistics, and Depth Anything V2 frozen.
AdamW uses weight decay $10^{-4}$ and learning rates $10^{-4}$ for refinement and gating and $10^{-5}$ for trainable initializer layers. The optimizer and learning-rate schedule are initialized separately for each stage, with learning rates halved every 10 epochs within each stage. Training uses FP32, batch size 16, and four NVIDIA RTX 4090 GPUs.




\begin{table}[t]
\centering
\caption{Quantitative comparison on TransCG. Accuracy is averaged per image across all 23,524 test frames at $320\times240$ following the official evaluation protocol.}
\footnotesize
\setlength{\tabcolsep}{2.7pt}
\begin{tabular}{@{}lrrrrrr@{}}
\toprule
Method & MAE$\downarrow$ & RMSE$\downarrow$ & REL$\downarrow$ & $\delta_{1.05}\!\uparrow$ & $\delta_{1.10}\!\uparrow$ & $\delta_{1.25}\!\uparrow$ \\
\midrule
DFNet~\cite{fang2022transcg} & 12.567 & 17.543 & 0.02567 & 84.94 & 96.57 & 99.85 \\
BP-Net~\cite{tang2024bilateral} & 37.855 & 61.796 & 0.07680 & 51.68 & 69.33 & 97.98 \\
PSD~\cite{chen2025propagating} & 37.693 & 59.735 & 0.07660 & 51.34 & 68.81 & 97.96 \\
ReMake~\cite{cheng2026rethinking} & 8.040 & 11.363 & 0.01660 & 93.63 & 98.65 & \textbf{99.99} \\
MIR-Depth (Ours) & \textbf{6.833} & \textbf{10.504} & \textbf{0.01397} & \textbf{94.41} & \textbf{98.76} & \textbf{99.99} \\
\bottomrule
\end{tabular}

\label{tab:transcg_results}
\end{table}

\begin{figure*}[t]
\centering
\includegraphics[width=0.816\textwidth]{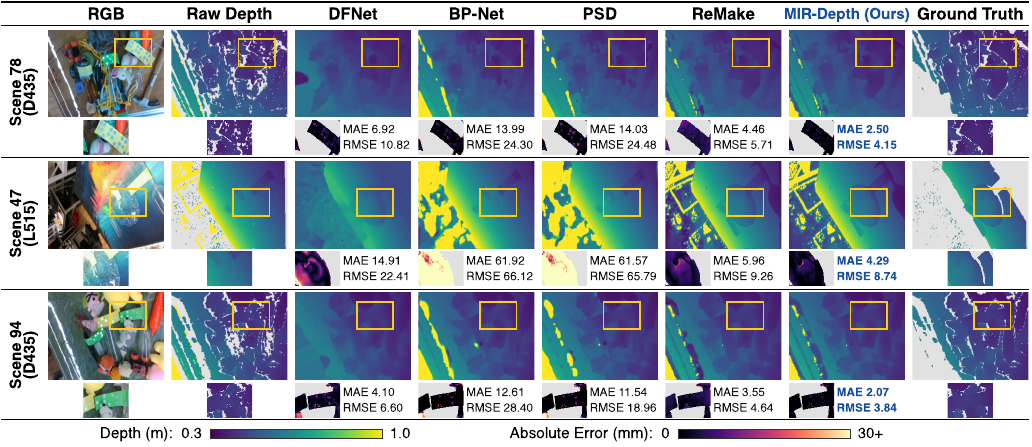}
\vspace{-0.35em}
\caption{\textbf{Qualitative comparison on TransCG.} Examples span D435 and L515 test scenes, where yellow boxes mark zoomed regions and bottom error maps visualize pixel-wise absolute error in millimeters on valid transparent-object pixels alongside full-frame ROI MAE and RMSE.}
\label{fig:transcg_qualitative}
\vspace{-0.35em}
\end{figure*}

\subsection{Evaluation on TransCG}
\label{sec:exp_transcg}

\begin{table*}[t]
\centering
\caption{Zero-shot cross-dataset evaluation. Depth errors in millimeters are aggregated pixel-wise over valid transparent-object pixels at native resolution without fine-tuning ($^{\ddagger}$132,596 valid PSD predictions on ClearPose).}
\footnotesize
\setlength{\tabcolsep}{6.0pt}
\begin{tabular*}{\textwidth}{@{\extracolsep{\fill}}lrrrrrrrr@{}}
\toprule
& \multicolumn{2}{c}{TODD (4,339)} & \multicolumn{2}{c}{TRansPose (32,734)} & \multicolumn{2}{c}{ClearPose (132,598)} & \multicolumn{2}{c}{HouseCat6D (2,929)} \\
\cmidrule(lr){2-3}\cmidrule(lr){4-5}\cmidrule(lr){6-7}\cmidrule(lr){8-9}
Method & MAE $\downarrow$ & RMSE $\downarrow$ & MAE $\downarrow$ & RMSE $\downarrow$ & MAE $\downarrow$ & RMSE $\downarrow$ & MAE $\downarrow$ & RMSE $\downarrow$ \\
\midrule
DFNet~\cite{fang2022transcg} & 69.459 & 101.206 & 116.499 & 138.225 & 127.729 & \textbf{224.717} & 45.048 & 64.050 \\
BP-Net~\cite{tang2024bilateral} & 87.924 & 1291.105 & 148.073 & 223.566 & 177.845 & 416.147 & 104.314 & 245.017 \\
PSD~\cite{chen2025propagating} & 88.464 & 1305.920 & 306.267 & 651.240 & 172.320$^{\ddagger}$ & 412.012$^{\ddagger}$ & 100.743 & 245.879 \\
ReMake~\cite{cheng2026rethinking} & 37.177 & 54.402 & 77.890 & 100.471 & 309.147 & 378.079 & 40.192 & 52.405 \\
MIR-Depth (Ours) & \textbf{32.915} & \textbf{49.797} & \textbf{72.591} & \textbf{95.128} & \textbf{122.294} & 238.050 & \textbf{36.429} & \textbf{48.035} \\
\bottomrule
\end{tabular*}

\label{tab:cross_dataset}
\end{table*}

\noindent\textbf{Setup and baselines.}
The TransCG test split contains 23,524 frames from 52 scenes (11,766 D435 and 11,758 L515 frames).
Predictions are bilinearly resized to $320\times240$ and evaluated on $\Omega_{\mathrm e}=\{p:M_{\mathrm e}(p)=1,\ V_{\mathrm e}(p)=1\}$, where $M_{\mathrm e}$ is the evaluation transparent-object mask, $V_{\mathrm e}$ is the validity mask retained before depth inpainting in the official preprocessing, and $D^*$ denotes the processed evaluation target.
Per-image MAE, RMSE, $\mathrm{REL}=\langle|\widehat D-D^*|/(D^*+\epsilon_{\mathrm e})\rangle_{\Omega_{\mathrm e}}$, and threshold accuracy $\delta_\tau$ are averaged across images.
The official implementation uses $\epsilon_{\mathrm e}=10^{-8}\,\mathrm{m}$ and counts pixels satisfying $\max(\widehat D/(D^*+\epsilon_{\mathrm e}),D^*/\widehat D)<\tau$ for $\tau\in\{1.05,1.10,1.25\}$, normalizing percentages by $|\Omega_{\mathrm e}|$.
For nonnegative predictions, zero predictions fail the ratio test; empty $\Omega_{\mathrm e}$ gives zero per-image errors and $100\%$ threshold accuracy.
The baselines cover three categories: DFNet~\cite{fang2022transcg} performs in-domain single-pass regression without external priors; BP-Net~\cite{tang2024bilateral} (NYUv2) and PSD~\cite{chen2025propagating} (NYUv2 and KITTI with Depth Anything V2) evaluate opaque-scene propagation under refraction; and ReMake~\cite{cheng2026rethinking} provides prior-guided single-pass prediction using the same Depth Anything V2 Small prior and mask inputs as MIR-Depth.

\noindent\textbf{Main results.}
Table~\ref{tab:transcg_results} shows that BP-Net and PSD propagate refracted background depth across foreground objects by treating nonzero sensor returns as reliable surface seeds, yielding $37.855/61.796\,\mathrm{mm}$ and $37.693/59.735\,\mathrm{mm}$ MAE/RMSE, respectively, even with Depth Anything V2 priors in PSD.
Trained on transparent scenes, DFNet reduces error to $12.567/17.543\,\mathrm{mm}$, but its single-pass CNN without monocular shape priors struggles with large specular holes.
ReMake combines the same Depth Anything V2 prior and object mask to reach $8.040/11.363\,\mathrm{mm}$, but its one-shot decoder cannot jointly resolve large interior refraction and millimeter-scale rim detail.
Using this prior-guided initial anchor, MIR-Depth redesigns single-view estimation through Coordinate-Aware Multi-Range Hypotheses, whose 3D-aggregated global volume captures large interior offsets, reducing MAE/RMSE to $6.833/10.504\,\mathrm{mm}$ and $\mathrm{REL}$ to $0.01397$, compared with $0.01660$ for ReMake.
Selective Recurrent Refinement fuses active-range lookup features for inverse-depth updates, with MIR-Depth achieving $\delta_{1.05}=94.41\%$ and $\delta_{1.10}=98.76\%$, compared with $93.63\%$ and $98.65\%$ for ReMake.
Fig.~\ref{fig:transcg_qualitative} further shows that global hypothesis aggregation pulls refracted vessel interiors back to the visible surface in D435 captures (Scenes~78 and 94, $2.50$ and $2.07\,\mathrm{mm}$ MAE versus $4.46$ and $3.55\,\mathrm{mm}$ for ReMake) and L515 captures (Scene~47, $4.29\,\mathrm{mm}$ versus $5.96\,\mathrm{mm}$), while depth-axis local filtering preserves sharp container rims.


\begin{table}[!t]
\centering
\caption{Hole filling and refraction correction on TransCG. Across 23,524 frames at $640\times480$, $19.06\%$ of ROI pixels are holes ($D_{\mathrm{raw}}=0$), while nonzero returns ($D_{\mathrm{raw}}>0$) have $32.121\,\mathrm{mm}$ raw MAE. Metrics include within-$20\,\mathrm{mm}$ hole accuracy (Hole Acc.), nonzero pixels modified by $>1\,\mathrm{mm}$ (Chg.), error-reduced nonzero pixels (Impr.), $>20\,\mathrm{mm}$ outliers corrected to within $20\,\mathrm{mm}$ (Repair), nonzero output MAE in millimeters (Out.\ MAE), and MAE reduction $r_{\mathrm{MAE}}$.}
\footnotesize
\setlength{\tabcolsep}{2.2pt}
\begin{tabular}{@{}lrrrrrr@{}}
\toprule
Method & Hole Acc.$\uparrow$ & Chg. & Impr.$\uparrow$ & Repair$\uparrow$ & Out.\ MAE$\downarrow$ & $r_{\mathrm{MAE}}\!\uparrow$ \\
\midrule
DFNet~\cite{fang2022transcg} & 63.26 & 88.36 & 51.71 & 41.46 & 22.511 & 29.92 \\
BP-Net~\cite{tang2024bilateral} & 48.45 & 21.36 & 32.93 & 0.80 & 32.297 & $-0.55$ \\
PSD~\cite{chen2025propagating} & 43.17 & 0.00 & 34.48 & 0.00 & 32.121 & 0.00 \\
ReMake~\cite{cheng2026rethinking} & 91.26 & 93.79 & 62.47 & 88.52 & 7.137 & 77.78 \\
MIR-Depth & \textbf{92.32} & 72.06 & \textbf{64.80} & \textbf{89.74} & \textbf{5.584} & \textbf{82.62} \\
\bottomrule
\end{tabular}


\label{tab:transcg_diag}
\vspace{-0.4em}
\end{table}

\noindent\textbf{Hole filling and refraction correction.}
Table~\ref{tab:transcg_diag} separately evaluates missing depth and nonzero returns.
Anchoring predictions to raw measurements leaves PSD modifying $0.00\%$ of nonzero pixels by more than $1\,\mathrm{mm}$, with MAE remaining at the raw value, while BP-Net repairs only $0.80\%$ of outliers.
MIR-Depth predicts positive depth at all evaluated holes, whereas ReMake leaves 33 nonpositive predictions; their within-$20\,\mathrm{mm}$ hole accuracies are $92.32\%$ and $91.26\%$, respectively.
Using the masked raw-depth residual $\delta_{\mathrm{obs}}$ as soft evidence, MIR-Depth repairs $89.74\%$ of $>20\,\mathrm{mm}$ outliers on nonzero returns and achieves $5.584\,\mathrm{mm}$ MAE ($82.62\%$ reduction), compared with $88.52\%$ repair and $7.137\,\mathrm{mm}$ MAE ($77.78\%$ reduction) for ReMake.
Relative to raw sensor depth, MIR-Depth modifies $72.06\%$ of nonzero pixels by more than $1\,\mathrm{mm}$ and reduces error at $64.80\%$.


\subsection{Cross-Dataset Evaluation}
\label{sec:exp_transfer}

To evaluate zero-shot generalization across sensors, object categories, and clutter densities, we compare MIR-Depth with general-purpose depth completion methods BP-Net (trained on NYUv2) and PSD (trained on NYUv2 and KITTI), and transparent-object completion methods DFNet and ReMake (both trained on TransCG).
Without fine-tuning, all five models are evaluated on TODD~\cite{xu2022seeingglass} (4,339 frames), TRansPose~\cite{kim2024transpose} (32,734 right-camera frames across 26 sequences), ClearPose~\cite{chen2022clearpose} (132,598 frames across labels 1--63), and HouseCat6D~\cite{jung2024housecat6d} (2,929 frames).
Errors are aggregated pixel-wise at native resolution over the intersection of transparent-object masks and valid ground-truth depth ($D^*\geq0.01\,\mathrm{m}$ on TODD, ClearPose, and HouseCat6D; CAD depth $D^*>0.01\,\mathrm{m}$ for categories C, H, and T on TRansPose).

Table~\ref{tab:cross_dataset} shows that opaque-scene propagation in BP-Net and PSD amplifies outliers under unseen refraction patterns, reaching $1291.105$ and $1305.920\,\mathrm{mm}$ RMSE on TODD and $306.267\,\mathrm{mm}$ MAE for PSD on TRansPose.
Despite sharing the Depth Anything V2 prior, ReMake's single-pass absolute-depth regression from domain-coupled features induces severe scale drift on ClearPose ($309.147/378.079\,\mathrm{mm}$ MAE/RMSE).
Through explicit candidate conditioning $c(q,p)$ in Coordinate-Aware Multi-Range Hypotheses, MIR-Depth evaluates inverse-depth candidates using the raw-depth residual $\delta_{\mathrm{obs}}$ normalized within $c(q,p)$ to re-anchor metric scale across camera domains, reducing ClearPose MAE/RMSE to $122.294/238.050\,\mathrm{mm}$ and achieving lower MAE than DFNet ($127.729\,\mathrm{mm}$).
Selective Recurrent Refinement masks out-of-bounds local lookups and re-weights valid intervals to fuse features from fixed candidate grids, reducing MAE/RMSE against ReMake on TODD ($32.915/49.797$ vs. $37.177/54.402\,\mathrm{mm}$), TRansPose ($72.591/95.128$ vs. $77.890/100.471\,\mathrm{mm}$), and HouseCat6D ($36.429/48.035$ vs. $40.192/52.405\,\mathrm{mm}$).

\begin{figure}[t]
\centering
\includegraphics[width=0.94\linewidth]{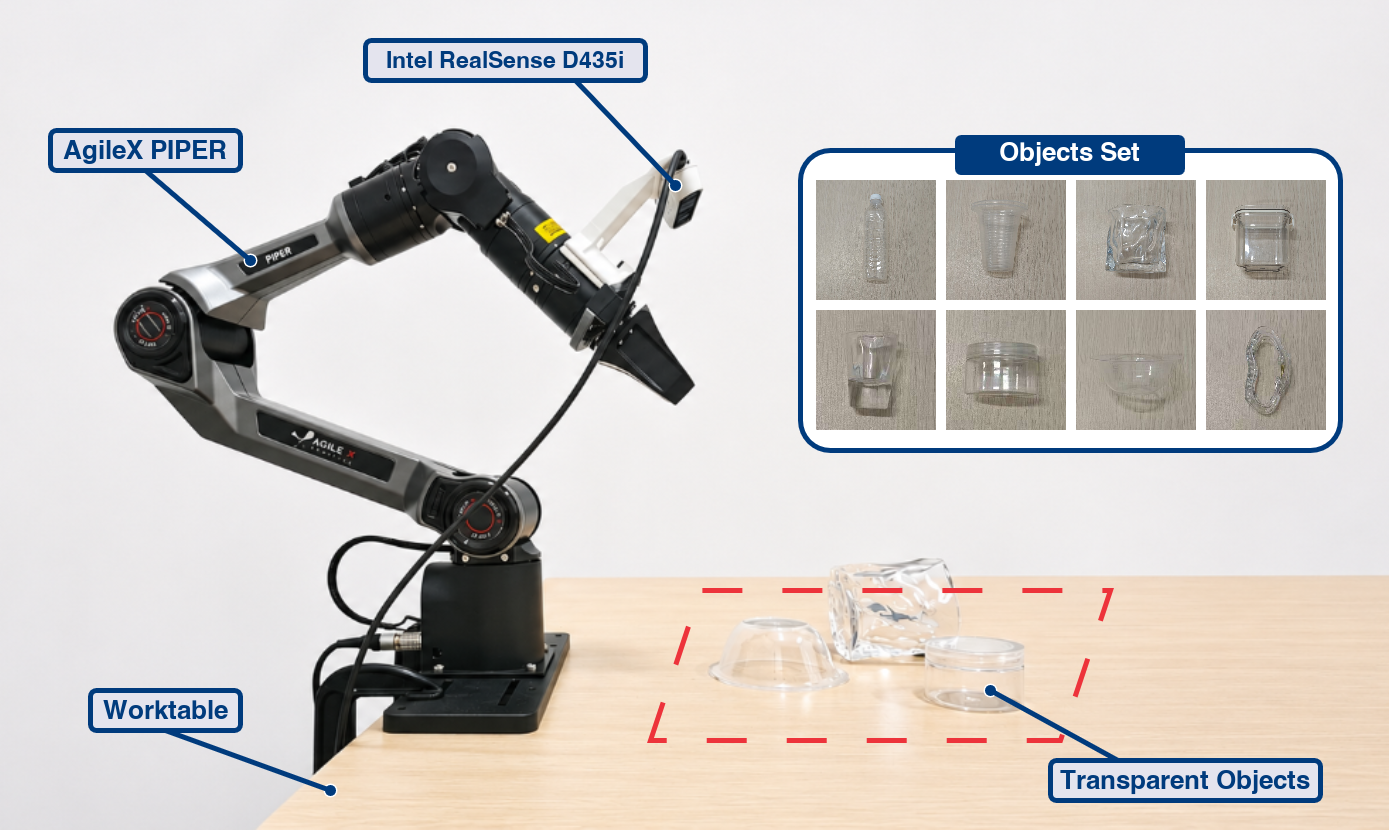}
\vspace{-0.3em}
\caption{\textbf{Real-world tabletop platform.} Our setup comprises a 6-DoF AgileX PIPER robot arm, a wrist-mounted Intel RealSense D435i camera, and eight household transparent objects shown in the inset.}
\label{fig:real_exp_setup}
\vspace{-0.4em}
\end{figure}

\begin{figure*}[t]
\centering
\includegraphics[width=0.95\textwidth]{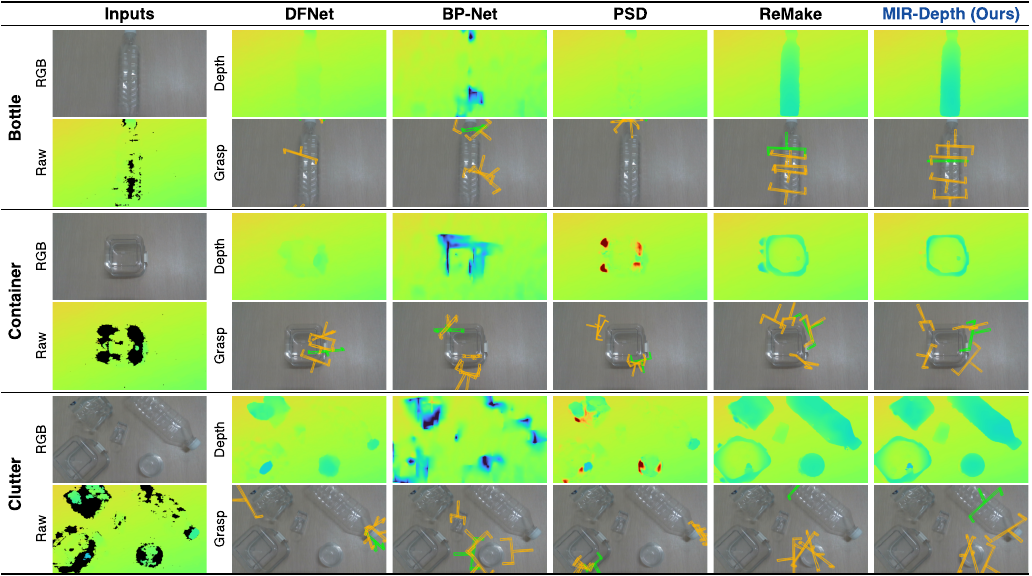}
\vspace{-0.35em}
\caption{\textbf{Real-world depth completion and 6-DoF grasp pose prediction.} Comparison across an upright bottle, a hollow container, and multi-object clutter captured on the platform in Fig.~\ref{fig:real_exp_setup}, where black pixels in \textit{Raw} denote missing sensor returns.}
\label{fig:real_world_depth}
\vspace{-0.5em}
\end{figure*}

\subsection{Real-World Evaluation}
\label{sec:real_world_qualitative}

\noindent\textbf{Real-world depth completion.}
We evaluate all models on our AgileX PIPER platform with a wrist-mounted RealSense D435i (Fig.~\ref{fig:real_exp_setup}), using $R=f_{\mathrm{DAv2-S}}(I)$ and masks $M=M_{\mathrm{SAM}}\in\{0,1\}^{1\times H\times W}$ generated by the Segment Anything Model (SAM)~\cite{kirillov2023segment}.
In Fig.~\ref{fig:real_world_depth}, DFNet blurs object silhouettes into the tabletop, while BP-Net and PSD propagate refracted background readings across hollow interiors.
ReMake recovers overall object presence but over-smooths thin container walls and merges adjacent transparent boundaries in clutter.
MIR-Depth recovers curved bottle walls, preserves sharper container boundaries, and reduces residual overshoot along imperfect SAM mask borders.

\noindent\textbf{6-DoF grasp pose prediction.}
Using a shared GraspNet-based 6-DoF grasp detector~\cite{fang2020graspnet} (the \textit{Grasp} rows in Fig.~\ref{fig:real_world_depth}), depth maps from DFNet, BP-Net, and PSD produce predicted grippers that collide with the tabletop or miss collapsed cavities.
MIR-Depth recovers crisp rim thickness and separated object surfaces, enabling clean antipodal jaw placement along curved bottle bodies, thin container rims, and cluttered transparent objects.

\subsection{Ablation Study}
\label{sec:exp_ablation}

\begin{table}[!t]
\centering
\caption{Ablation study on TransCG.
All variants are evaluated on 23,524 test frames at $320\times240$.
MAE and RMSE are reported in millimeters, and $\delta_{1.05}$ in percent.
Underlined values indicate the best results; boldface highlights the full model.}
\footnotesize
\setlength{\tabcolsep}{4.8pt}
\begin{tabular}{@{}lrrrr@{}}
\toprule
Variant & MAE$\downarrow$ & RMSE$\downarrow$ & REL$\downarrow$ & $\delta_{1.05}\!\uparrow$ \\
\midrule
w/o Iterative refinement & 8.040 & 11.363 & 0.01660 & 93.63 \\
w/o Local ranges & 6.895 & 10.540 & 0.01412 & 94.30 \\
w/o Selective fusion & 6.906 & 10.597 & 0.01412 & 94.17 \\
w/o Sequence loss & 6.882 & 10.554 & 0.01409 & 94.29 \\
w/o gating  & \underline{6.810} & \underline{10.245} & \underline{0.01396} & \underline{94.45} \\
MIR-Depth (Ours) & \textbf{6.833} & \textbf{10.504} & \textbf{0.01397} & \textbf{94.41} \\
\bottomrule
\end{tabular}

\label{tab:ablation}
\vspace{-0.4em}
\end{table}

\noindent\textbf{Component ablations.}
Table~\ref{tab:ablation} compares MIR-Depth with five ablated variants on 23,524 TransCG test frames at $320\times240$.
Removing iterative refinement increases MAE/RMSE from $6.833/10.504$ to $8.040/11.363\,\mathrm{mm}$, showing the contribution of recurrent candidate lookup and depth updates to correcting initial errors.
Replacing the medium and fine candidate grids with the global grid, while retaining all branches, aggregation operators, and per-branch candidate counts, increases MAE/RMSE to $6.895/10.540\,\mathrm{mm}$ and reduces $\delta_{1.05}$ from $94.41\%$ to $94.30\%$, supporting local ranges at a fixed candidate count.
Replacing learned selection with uniform weights over active ranges, while preserving candidate grids and lookup validity masking, increases MAE/RMSE to $6.906/10.597\,\mathrm{mm}$ and reduces $\delta_{1.05}$ to $94.17\%$, showing the benefit of weighting ranges using lookup features, entropy, and peak probability.
Removing intermediate supervision increases MAE/RMSE to $6.882/10.554\,\mathrm{mm}$, supporting supervision of intermediate refinement outputs.

The \emph{w/o gating} variant achieves the best TransCG results ($6.810/10.245\,\mathrm{mm}$ MAE/RMSE, $0.01396$ REL, and $94.45\%$ $\delta_{1.05}$).
Compared with full MIR-Depth, MAE and RMSE decrease by $0.023\,\mathrm{mm}$ ($0.34\%$) and $0.259\,\mathrm{mm}$ ($2.47\%$), while $\delta_{1.05}$ increases by $0.04$ percentage points.
Setting $G\equiv1$ directly outputs $D^T$, whereas $G<1$ retains part of $D_0$ (Eq.~\ref{eq:gated_prediction}). In the ungated variant, gate parameters are frozen and the correction regularizer remains active on $D^T-D_0$.
When $D^T$ is accurate, residual attenuation can retain initialization errors, which may contribute to the advantage of the ungated variant on TransCG.

\begin{figure}[t]
\centering
\includegraphics[width=0.93\linewidth]{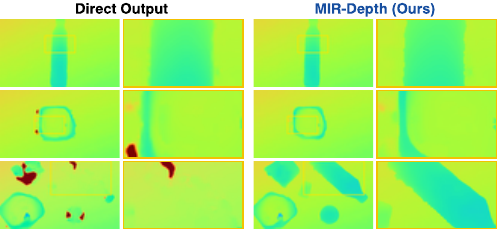}
\vspace{-0.3em}

\caption{\textbf{Spatial-gating ablation in real scenes.} The \emph{w/o gating} variant (\textit{Direct Output}, \textit{left}), which sets $G\equiv1$ and directly outputs $D^T$, is compared with full \textbf{MIR-Depth (Ours)} (\textit{right}) on the bottle, container, and clutter scenes (top to bottom) from Fig.~\ref{fig:real_world_depth}. In the zoomed yellow boxes, full MIR-Depth exhibits fewer boundary depth jumps and less surface fragmentation than the ungated variant.}

\label{fig:real_scene_gating_ablation}
\vspace{-0.4em}
\end{figure}

In real-world captures with predicted SAM masks (Fig.~\ref{fig:real_scene_gating_ablation}), ungated output produces abrupt depth changes near the hollow container and fragmented surfaces in clutter, whereas full MIR-Depth preserves more continuous contours.
Spatially Gated Correction uses recurrent features and candidate-distribution cues to attenuate the residual, reducing boundary depth jumps and surface fragmentation.

\section{Conclusion}
\label{sec:conclusion}

We present MIR-Depth for single-view RGB-D depth completion of transparent objects through prior-guided initial depth refinement. Coordinate-Aware Multi-Range Hypotheses combine shared global and pixel-wise local volumes to capture large refraction offsets and fine geometry. Selective Recurrent Refinement updates inverse depth through valid-range lookup, while Spatially Gated Correction preserves accurate initial regions. Experiments on TransCG, four zero-shot benchmarks, and a real robotic grasping platform show that MIR-Depth achieves the best results among compared baselines on multiple transparent-object datasets.

\noindent\textbf{Limitations.}
MIR-Depth recovers visible surfaces from a single RGB-D frame despite mostly missing or refraction-corrupted sensor measurements, but cannot reveal unobserved geometry. A fully transparent object against a uniform background may have almost no visible boundaries, with its far side hidden; even a human cannot determine its exact shape from that frame. Future work will extend MIR-Depth to temporal RGB-D streams, using views captured during robot arm motion to reveal hidden surfaces and enable closed-loop manipulation of transparent objects.



\bibliographystyle{IEEEtran}
\bibliography{main}

\begin{thebibliography}{10}
\providecommand{\url}[1]{#1}
\csname url@samestyle\endcsname
\providecommand{\newblock}{\relax}
\providecommand{\bibinfo}[2]{#2}
\providecommand{\BIBentrySTDinterwordspacing}{\spaceskip=0pt\relax}
\providecommand{\BIBentryALTinterwordstretchfactor}{4}
\providecommand{\BIBentryALTinterwordspacing}{\spaceskip=\fontdimen2\font plus
\BIBentryALTinterwordstretchfactor\fontdimen3\font minus \fontdimen4\font\relax}
\providecommand{\BIBforeignlanguage}[2]{{%
\expandafter\ifx\csname l@#1\endcsname\relax
\typeout{** WARNING: IEEEtran.bst: No hyphenation pattern has been}%
\typeout{** loaded for the language `#1'. Using the pattern for}%
\typeout{** the default language instead.}%
\else
\language=\csname l@#1\endcsname
\fi
#2}}
\providecommand{\BIBdecl}{\relax}
\BIBdecl

\bibitem{fang2020graspnet}
H.-S. Fang, C.~Wang, M.~Gou, and C.~Lu, ``Graspnet-1billion: A large-scale benchmark for general object grasping,'' in \emph{2020 IEEE/CVF Conference on Computer Vision and Pattern Recognition (CVPR)}.\hskip 1em plus 0.5em minus 0.4em\relax IEEE, 2020, pp. 11\,441--11\,450.

\bibitem{sajjan2020clear}
S.~Sajjan, M.~Moore, M.~Pan, G.~Nagaraja, J.~Lee, A.~Zeng, and S.~Song, ``Clear grasp: 3d shape estimation of transparent objects for manipulation,'' in \emph{2020 IEEE international conference on robotics and automation (ICRA)}.\hskip 1em plus 0.5em minus 0.4em\relax IEEE, 2020, pp. 3634--3642.

\bibitem{liu2020keypose}
X.~Liu, R.~Jonschkowski, A.~Angelova, and K.~Konolige, ``Keypose: Multi-view 3d labeling and keypoint estimation for transparent objects,'' in \emph{2020 IEEE/CVF Conference on Computer Vision and Pattern Recognition (CVPR)}.\hskip 1em plus 0.5em minus 0.4em\relax IEEE, 2020, pp. 11\,599--11\,607.

\bibitem{zhu2021rgb}
L.~Zhu, A.~Mousavian, Y.~Xiang, H.~Mazhar, J.~van Eenbergen, S.~Debnath, and D.~Fox, ``Rgb-d local implicit function for depth completion of transparent objects,'' \emph{arXiv preprint arXiv:2104.00622}, 2021.

\bibitem{cheng2018depth}
X.~Cheng, P.~Wang, and R.~Yang, ``Depth estimation via affinity learned with convolutional spatial propagation network,'' in \emph{European Conference on Computer Vision}.\hskip 1em plus 0.5em minus 0.4em\relax Springer, 2018, pp. 108--125.

\bibitem{park2020non}
J.~Park, K.~Joo, Z.~Hu, C.-K. Liu, and I.~So~Kweon, ``Non-local spatial propagation network for depth completion,'' in \emph{European conference on computer vision}.\hskip 1em plus 0.5em minus 0.4em\relax Springer, 2020, pp. 120--136.

\bibitem{zhang2023completionformer}
Y.~Zhang, X.~Guo, M.~Poggi, Z.~Zhu, G.~Huang, and S.~Mattoccia, ``Completionformer: Depth completion with convolutions and vision transformers,'' in \emph{2023 IEEE/CVF Conference on Computer Vision and Pattern Recognition (CVPR)}.\hskip 1em plus 0.5em minus 0.4em\relax IEEE, 2023, pp. 18\,527--18\,536.

\bibitem{cheng2026rethinking}
Y.~Cheng, X.~Gao, S.~Zhang, C.~Zeng, F.~Zha, L.~Sun, and C.~Yang, ``Rethinking transparent object grasping: Depth completion with monocular depth estimation and instance mask,'' \emph{IEEE Robotics and Automation Letters}, 2026.

\bibitem{fang2022transcg}
H.~Fang, H.-S. Fang, S.~Xu, and C.~Lu, ``Transcg: A large-scale real-world dataset for transparent object depth completion and a grasping baseline,'' \emph{IEEE Robotics and Automation Letters}, vol.~7, no.~3, pp. 7383--7390, 2022.

\bibitem{chen2022tode}
K.~Chen, S.~Wang, B.~Xia, D.~Li, Z.~Kan, and B.~Li, ``Tode-trans: Transparent object depth estimation with transformer,'' \emph{arXiv preprint arXiv:2209.08455}, 2022.

\bibitem{li2023fdct}
T.~Li, Z.~Chen, H.~Liu, and C.~Wang, ``Fdct: Fast depth completion for transparent objects,'' \emph{IEEE Robotics and Automation Letters}, vol.~8, no.~9, pp. 5823--5830, 2023.

\bibitem{ranftl2020towards}
R.~Ranftl, K.~Lasinger, D.~Hafner, K.~Schindler, and V.~Koltun, ``Towards robust monocular depth estimation: Mixing datasets for zero-shot cross-dataset transfer,'' \emph{IEEE transactions on pattern analysis and machine intelligence}, vol.~44, no.~3, pp. 1623--1637, 2020.

\bibitem{yang2024depth}
L.~Yang, B.~Kang, Z.~Huang, X.~Xu, J.~Feng, and H.~Zhao, ``Depth anything: Unleashing the power of large-scale unlabeled data,'' in \emph{2024 IEEE/CVF Conference on Computer Vision and Pattern Recognition (CVPR)}.\hskip 1em plus 0.5em minus 0.4em\relax IEEE, 2024, pp. 10\,371--10\,381.

\bibitem{yang2024depthv2}
L.~Yang, B.~Kang, Z.~Huang, Z.~Zhao, X.~Xu, J.~Feng, and H.~Zhao, ``Depth anything v2,'' \emph{Advances in neural information processing systems}, vol.~37, pp. 21\,875--21\,911, 2024.

\bibitem{gu2020cascade}
X.~Gu, Z.~Fan, S.~Zhu, Z.~Dai, F.~Tan, and P.~Tan, ``Cascade cost volume for high-resolution multi-view stereo and stereo matching,'' in \emph{2020 IEEE/CVF Conference on Computer Vision and Pattern Recognition (CVPR)}.\hskip 1em plus 0.5em minus 0.4em\relax IEEE, 2020, pp. 2492--2501.

\bibitem{xu2025igevpp}
G.~Xu, X.~Wang, Z.~Zhang, J.~Cheng, C.~Liao, and X.~Yang, ``Igev++: Iterative multi-range geometry encoding volumes for stereo matching,'' \emph{IEEE Transactions on Pattern Analysis and Machine Intelligence}, vol.~47, no.~8, pp. 7108--7122, 2025.

\bibitem{tang2024bilateral}
J.~Tang, F.-P. Tian, B.~An, J.~Li, and P.~Tan, ``Bilateral propagation network for depth completion,'' in \emph{2024 IEEE/CVF Conference on Computer Vision and Pattern Recognition (CVPR)}.\hskip 1em plus 0.5em minus 0.4em\relax IEEE, 2024, pp. 9763--9772.

\bibitem{chen2025propagating}
S.~Chen, X.~Ma, H.~Zhang, H.~Li, and Z.~Wang, ``Propagating sparse depth via depth foundation model for out-of-distribution depth completion,'' \emph{IEEE Transactions on Image Processing}, 2025.

\bibitem{xu2022seeingglass}
\BIBentryALTinterwordspacing
H.~Xu, Y.~R. Wang, S.~Eppel, A.~Aspuru-Guzik, F.~Shkurti, and A.~Garg, ``Seeing glass: Joint point-cloud and depth completion for transparent objects,'' in \emph{Proceedings of the 5th Conference on Robot Learning}, ser. Proceedings of Machine Learning Research, A.~Faust, D.~Hsu, and G.~Neumann, Eds., vol. 164.\hskip 1em plus 0.5em minus 0.4em\relax PMLR, 08--11 Nov 2022, pp. 827--838. [Online]. Available: \url{https://proceedings.mlr.press/v164/xu22b.html}
\BIBentrySTDinterwordspacing

\bibitem{dai2022domain}
Q.~Dai, J.~Zhang, Q.~Li, T.~Wu, H.~Dong, Z.~Liu, P.~Tan, and H.~Wang, ``Domain randomization-enhanced depth simulation and restoration for perceiving and grasping specular and transparent objects,'' in \emph{European Conference on Computer Vision}.\hskip 1em plus 0.5em minus 0.4em\relax Springer, 2022, pp. 374--391.

\bibitem{hong2022cluedepth}
Y.~Hong, J.~Chen, Y.~Cheng, Y.~Han, F.~V. Reeth, L.~Claesen, and W.~Liu, ``Cluedepth grasp: Leveraging positional clues of depth for completing depth of transparent objects,'' \emph{Frontiers in Neurorobotics}, vol.~16, p. 1041702, 2022.

\bibitem{chen2022clearpose}
X.~Chen, H.~Zhang, Z.~Yu, A.~Opipari, and O.~Chadwicke~Jenkins, ``Clearpose: Large-scale transparent object dataset and benchmark,'' in \emph{European conference on computer vision}.\hskip 1em plus 0.5em minus 0.4em\relax Springer, 2022, pp. 381--396.

\bibitem{kim2024transpose}
J.~Kim, M.-H. Jeon, S.~Jung, W.~Yang, M.~Jung, J.~Shin, and A.~Kim, ``Transpose: Large-scale multispectral dataset for transparent object,'' \emph{The International Journal of Robotics Research}, vol.~43, no.~6, pp. 731--738, 2024.

\bibitem{jung2024housecat6d}
H.~Jung, S.-C. Wu, P.~Ruhkamp, G.~Zhai, H.~Schieber, G.~Rizzoli, P.~Wang, H.~Zhao, L.~Garattoni, S.~Meier \emph{et~al.}, ``Housecat6d-a large-scale multi-modal category level 6d object perception dataset with household objects in realistic scenarios,'' in \emph{Proceedings of the IEEE/CVF Conference on Computer Vision and Pattern Recognition}, 2024, pp. 22\,498--22\,508.

\bibitem{shi2024asgrasp}
J.~Shi, A.~Yong, Y.~Jin, D.~Li, H.~Niu, Z.~Jin, and H.~Wang, ``Asgrasp: Generalizable transparent object reconstruction and 6-dof grasp detection from rgb-d active stereo camera,'' in \emph{2024 IEEE international conference on robotics and automation (ICRA)}.\hskip 1em plus 0.5em minus 0.4em\relax IEEE, 2024, pp. 5441--5447.

\bibitem{min2026depthfocus}
J.~Min, J.~Kim, M.~Kim, C.-H. Min, Y.~Jeon, and M.~Choi, ``Depthfocus: Controllable depth estimation for see-through scenes,'' in \emph{Proceedings of the IEEE/CVF Conference on Computer Vision and Pattern Recognition (CVPR)}, June 2026, pp. 12\,595--12\,605.

\bibitem{wen2026seegroup}
H.~Wen and J.~Deng, ``Seegroup: Multi-layer depth estimation of transparent surfaces via self-determined grouping,'' in \emph{Proceedings of the IEEE/CVF Conference on Computer Vision and Pattern Recognition (CVPR)}, June 2026, pp. 7299--7309.

\bibitem{chang2018pyramid}
J.-R. Chang and Y.-S. Chen, ``Pyramid stereo matching network,'' in \emph{2018 IEEE/CVF conference on computer vision and pattern recognition}.\hskip 1em plus 0.5em minus 0.4em\relax IEEE, 2018, pp. 5410--5418.

\bibitem{teed2020raft}
Z.~Teed and J.~Deng, ``Raft: Recurrent all-pairs field transforms for optical flow,'' in \emph{European conference on computer vision}.\hskip 1em plus 0.5em minus 0.4em\relax Springer, 2020, pp. 402--419.

\bibitem{lipson2021raft}
L.~Lipson, Z.~Teed, and J.~Deng, ``Raft-stereo: Multilevel recurrent field transforms for stereo matching,'' in \emph{2021 International conference on 3D vision (3DV)}.\hskip 1em plus 0.5em minus 0.4em\relax IEEE, 2021, pp. 218--227.

\bibitem{xu2023iterative}
G.~Xu, X.~Wang, X.~Ding, and X.~Yang, ``Iterative geometry encoding volume for stereo matching,'' in \emph{2023 IEEE/CVF Conference on Computer Vision and Pattern Recognition (CVPR)}.\hskip 1em plus 0.5em minus 0.4em\relax IEEE, 2023, pp. 21\,919--21\,928.

\bibitem{shao2023iebins}
\BIBentryALTinterwordspacing
S.~Shao, Z.~Pei, X.~Wu, Z.~Liu, W.~Chen, and Z.~Li, ``Iebins: Iterative elastic bins for monocular depth estimation,'' in \emph{Advances in Neural Information Processing Systems}, A.~Oh, T.~Naumann, A.~Globerson, K.~Saenko, M.~Hardt, and S.~Levine, Eds., vol.~36.\hskip 1em plus 0.5em minus 0.4em\relax Curran Associates, Inc., 2023, pp. 53\,025--53\,037. [Online]. Available: \url{https://proceedings.neurips.cc/paper_files/paper/2023/file/a61023ce36d21010f1423304f8ec49af-Paper-Conference.pdf}
\BIBentrySTDinterwordspacing

\bibitem{xie2026dcirnet}
G.~Xie, Z.~Jiang, M.~Li, Y.~Zhang, Y.~Liu, Z.~Xie, and B.~Cao, ``Dcirnet: Depth completion with iterative refinement for dexterous grasping of transparent and reflective objects,'' \emph{Neurocomputing}, p. 134792, 2026.

\bibitem{wang2025transdiff}
H.~Wang, K.~Zhou, B.~Gu, Z.~Feng, W.~Wang, P.~Sun, Y.~Xiao, J.~Zhang, and H.~Dong, ``Transdiff: Diffusion-based method for manipulating transparent objects using a single rgb-d image,'' in \emph{2025 IEEE International Conference on Robotics and Automation (ICRA)}.\hskip 1em plus 0.5em minus 0.4em\relax IEEE, 2025, pp. 7277--7283.

\bibitem{kirillov2023segment}
A.~Kirillov, E.~Mintun, N.~Ravi, H.~Mao, C.~Rolland, L.~Gustafson, T.~Xiao, S.~Whitehead, A.~C. Berg, W.-Y. Lo, P.~Dollar, and R.~Girshick, ``Segment anything,'' in \emph{Proceedings of the IEEE/CVF International Conference on Computer Vision (ICCV)}, October 2023, pp. 4015--4026.

\end{thebibliography}

\end{document}